\documentclass [letterpaper]{IEEEtran}
\usepackage{listings}
\usepackage{xcolor}
\usepackage{tabularx}
\usepackage{comment}
\definecolor{codegreen}{rgb}{0,0.6,0}
\definecolor{codegray}{rgb}{0.5,0.5,0.5}
\definecolor{codepurple}{rgb}{0.58,0,0.82}
\definecolor{backcolour}{rgb}{0.95,0.95,0.92}
\lstdefinestyle{mystyle}{
  backgroundcolor=\color{backcolour}, commentstyle=\color{codegreen},
  keywordstyle=\color{magenta},
  numberstyle=\tiny\color{codegray},
  stringstyle=\color{codepurple},
  basicstyle=\ttfamily\footnotesize,
  breakatwhitespace=false,         
  breaklines=true,                 
  captionpos=b,                    
  keepspaces=true,                 
  numbers=left,                    
  numbersep=5pt,                  
  showspaces=false,                
  showstringspaces=false,
  showtabs=false,                  
  tabsize=2
}
\usepackage{graphicx,
	psfrag,
	epsfig,
	epstopdf,
	amsthm,
	amssymb,
	url,
	subcaption,
	balance,
	enumerate,
	color,
	setspace,
	tikz,
	pgfplots
}
\usepackage{amsmath}
\usepackage[nospace,noadjust]{cite}
\usepackage{pbox}
\usepackage{multirow}
\usepackage{adjustbox}
\usepackage{array}
\usepackage{booktabs}
\usepackage{tabularx}
\usepackage{float}
\usepackage{breqn}
\usepackage{algorithm}
\usepackage{algpseudocode}
\usepackage{amsmath}
\usetikzlibrary{shapes.geometric}

\newcommand{\cref}[1]{Constraint~\ref{#1}}

\newcommand{\ignore}[1]{}
\usepackage[paperwidth=8.5in, paperheight=11in,top=1.1in, bottom=0.996in, left=0.575in, right=0.6in]{geometry}

\begin{document}


\title{Closing the Responsibility Gap in AI-based Network Management: An Intelligent Audit System Approach }
	\author{
	\IEEEauthorblockN{ Emanuel Figetakis\IEEEauthorrefmark{1}, and Ahmed Refaey \IEEEauthorrefmark{1}}\\

	\IEEEauthorblockA{\IEEEauthorrefmark{1} University of Guelph, Guelph, Ontario, Canada.}}

\maketitle
\begin{abstract}

Existing network paradigms have achieved lower
downtime as well as a higher Quality of Experience (QoE) through
the use of Artificial Intelligence (AI)-based network management
tools. These AI management systems, allow for automatic responses
to changes in network conditions, lowering operation costs for
operators, and improving overall performance. While adopting
AI-based management tools enhance the overall network performance, it also introduce challenges such as removing human
supervision, privacy violations, algorithmic bias, and model inaccuracies. Furthermore, AI-based agents that fail to address these
challenges should be culpable themselves rather than the network
as a whole. To address this accountability gap, a framework
consisting of a Deep Reinforcement Learning (DRL) model and a
Machine Learning (ML) model is proposed to identify and assign
numerical values of responsibility to the AI-based management
agents involved in any decision-making regarding the network conditions,
which eventually affects the end-user. A simulation environment was
created for the framework to be trained using simulated network
operation parameters. The DRL model had a 96\% accuracy during testing for identifying the AI-based management agents, while the ML model using gradient descent learned the network conditions at an 83\% accuracy during testing. 
\end{abstract}


\section{Introduction}
Since the last decade, cellular network generations have evolved greatly utilizing the virtualized Evolved Packet Core (vEPC) with its ability to host and virtualize many of the network's functions. This method can integrate less physical hardware provided that virtualized functions now comprise the core  \cite{3gpp_2022}. Nevertheless, the decrease in hardware given the virtualization of the core has not solved the existing constraint of hardware limitations. Due to this constraint scalability is difficult without upgrading the network's infrastructure \cite{cisco}. Different methods try and increase the efficiency of the finite amount of computational resources given by a network's infrastructure such as Network Slicing (NS), however, this still is limited \cite{3gpp_NSM}. The authors of \cite{cisco} propose a new method adopted within the industry that grants a multi-vendor service space that does not rely on a central unit of hardware for the network. Through the disaggregation and decomposition of virtual functions in the core as well as the remote radio head (RRH), into separate cloudlets, a scalable solution is found. The multivendor network is harmonized with common feature sets across all target markets, allowing collaboration of the disaggregated virtualized network functions hosted on the cloud. While the system is highly adaptable to demands and outages and ease of management and orchestration of the specific virtualized network functions \cite{OpenSource-5G}, it introduces a multi-operator managed network. 


Each operator will employ the most efficient management systems and as a result of advancements made in Artificial Intelligence (AI), many cost-effective management systems currently utilize AI. The concept of utilizing AI for network orchestration and management is Zero Touch Networks (ZTN) and focuses on limiting human interaction with a network within the control domain. Virtualization of many network functions has also introduced the ability to virtualization AI tools within the Management and orchestration (MANO) domain service. The MANO service can orchestrate Network Function Virtualizations (NFVs), create and control Virtual Network Functions (VNFs), and manage the physical hardware that the service is being hosted on. The AI tools can be data-driven to make informed decisions on the network making them self-adaptive and self-reactive \cite{ZTN-1, ML-Healing, ML-Enabled}.

Admittedly there exists a gap in responsibility within AI-based utilized tools, making its inclusion in a multi-operator managed network alarming. Currently, there is no framework to address a multi-operator managed network, to not only identify changes made to the network by a specific operator but also to ensure fair and ethical use of the AI-based tools. This paper proposes a solution to this gap in responsibility by implementing an ease-of-use framework that can be placed in the elements of the vEPC which would support both legacy networks as well as virtualized networks. Along with the ease of use, it helps identify changes within the network as well as ensures ethical use of AI-based management tools which in turn becomes compliant with many new AI government regulations \cite{AIGOV-EU, AIGOV-CAD, AIGOV-USA}. 

In this work, a comprehensive system model is introduced for where the proposed framework will be hosted on the core network. The framework is designed to identify and find the AI-based management agents modifying the core network using network status logs. In addition, a method to measure the impact of automated AI-based tools on network
changes. The paper organization is as follows, Section \ref{SysModel} proposes the system model and problem definition, followed by Section \ref{Exper} which introduces experimentation and methods, then Section \ref{ReA} which discusses results and analysis, and finally Section \ref{Conc} which addresses future directions and concluding remarks. 

\begin{figure*}[h]
    \centering
    \includegraphics[width=\textwidth,height=70mm]{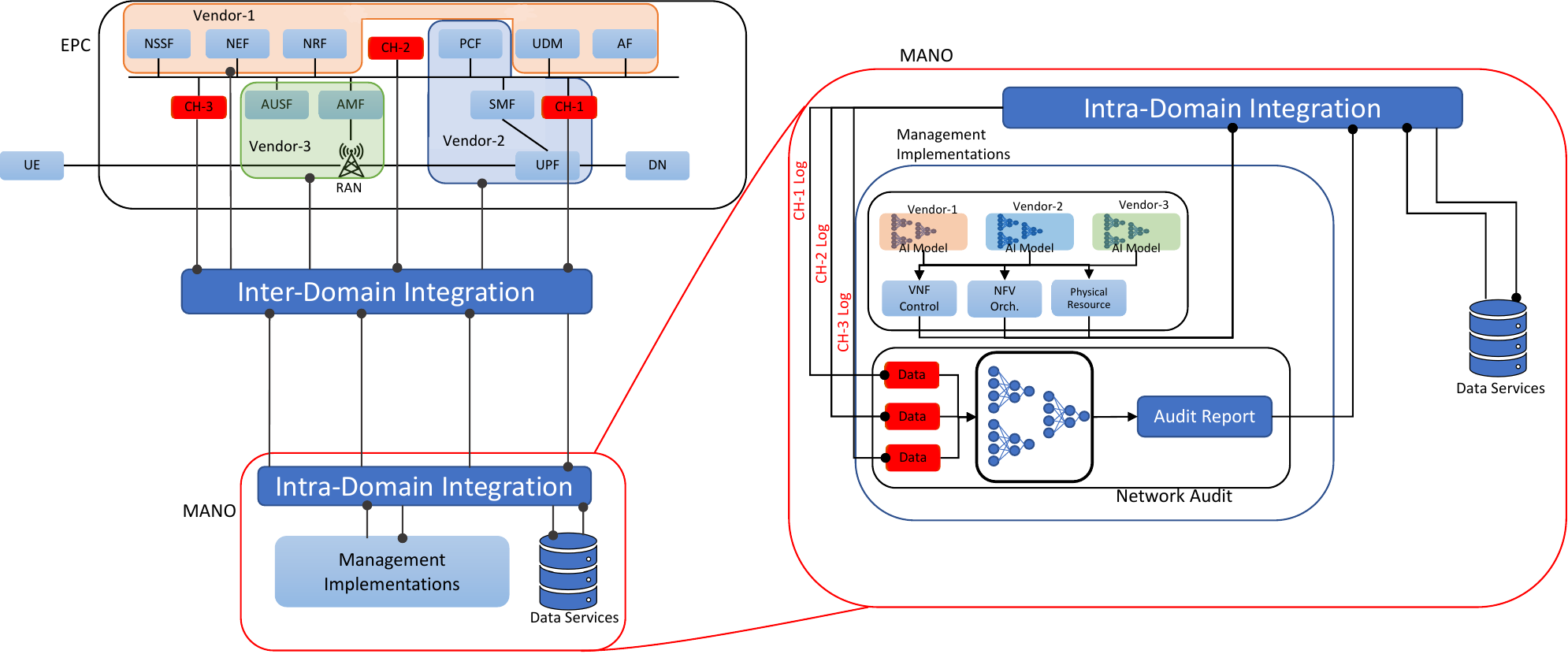}
    \caption{Proposed Network Architecture for Multi-Vendor Space with Management Plane.}
    \label{fig:PM}
\end{figure*}

\section{System Model}\label{SysModel}

Shown in Figure \ref{fig:PM} is the standard architecture of a ZTN \cite{3gpp_2022,etsi_2019}, a virtualized core powers the model's service network and has a separate domain for management, the MANO. The inter-domain integration handles the communication between both domains allowing for the transfer of network data. In the proposed model the core network adds passive nodes between known channels of communication between the vendors. These passive nodes collect the data from each vendor and forward it to the management plane this allows each of the vendors to access network information and make use of it in real-time with AI-based automation tools. 

\begin{figure}[htb]
    \centering
    \includegraphics[width=.48\textwidth,height=69mm]{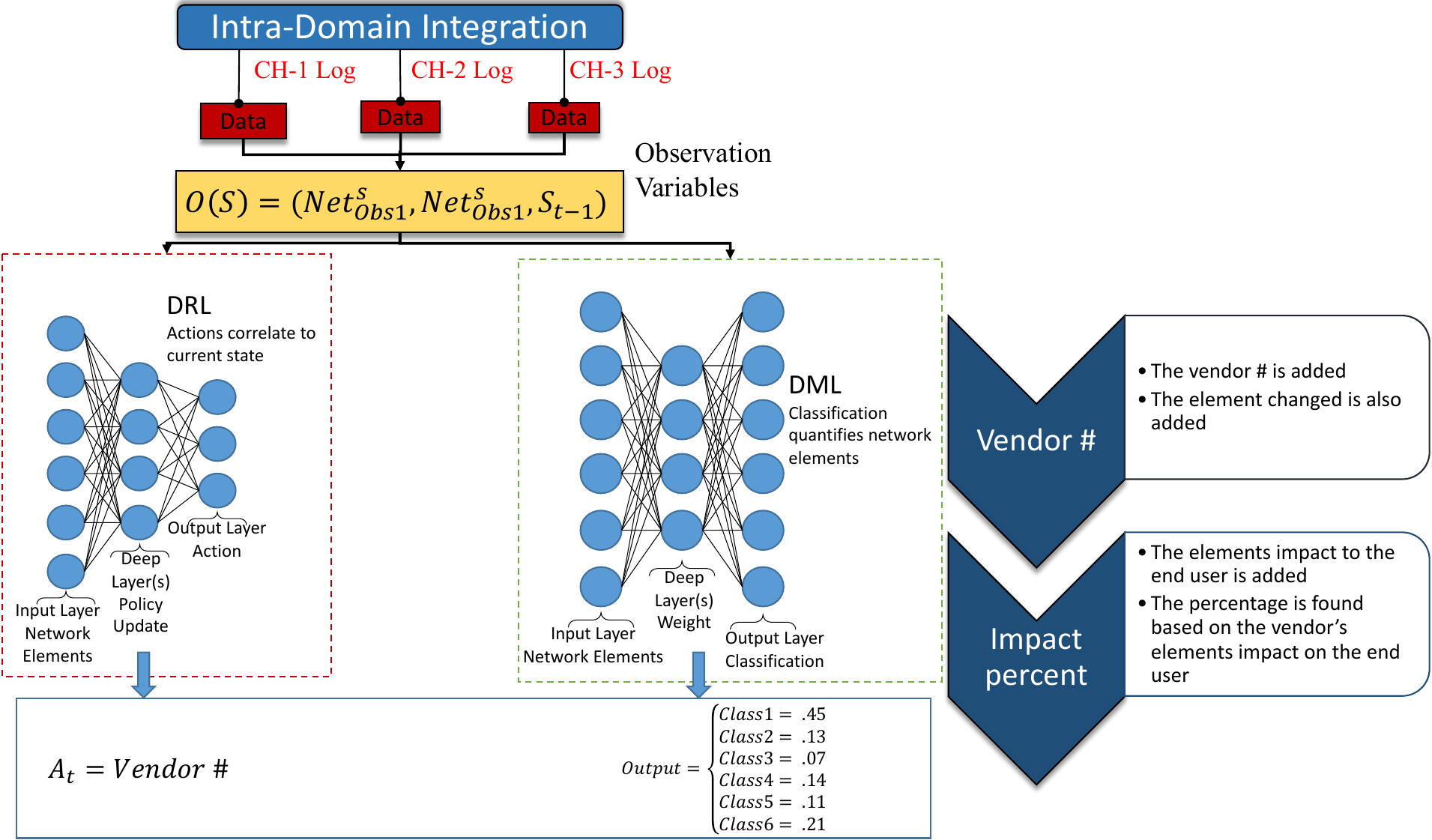}
    \caption{Intelligent Audit System.}
    \label{fig:MLFeature}
\end{figure}

The information gathered from the passive nodes will be used in the audit system later on. The nodes are collecting information about the network elements themselves. 
Given the system setup, the audit agent has only access to the change in the network elements. This makes learning possible only a Partially Observed Markov Decision Process (POMDP) since observation variables are only shared. However, this agent can only classify which management agent has changed the network, it cannot assign a percentage to each agent who is responsible for the impact on the end user. For this, a machine learning algorithm can be used, by using a multi-class classification method, the final layer of softmax activation functions, and the network elements as input. By leveraging the weights created from the learning of the machine learning algorithm the feature importance can be determined and responsibility can be determined this way. Figure \ref{fig:MLFeature} shows a comprehensive audit report can be created by combining this with the RL agent. 

\subsection{Problem Definition}\label{PD}

Given the system model and the goal of the framework, the problem definition began as follows, network elements have an effect on the end user, and the agents have an effect on the network elements. Each of the elements is within different domains of the network architecture, yet still influences the others. This formulation follows what is addressed in Graph Ranking \cite{Graph_Ranking}. The following section covers in detail the definition of the problem using Graph Ranking and solving through POMDP. The variables that are used in the problem are in Table \ref{tab:Vars}.

\subsection{Modeling through Graph Ranking}
\begin{figure}[htb]
    \centering
    \includegraphics[width=.45\textwidth,height=50mm]{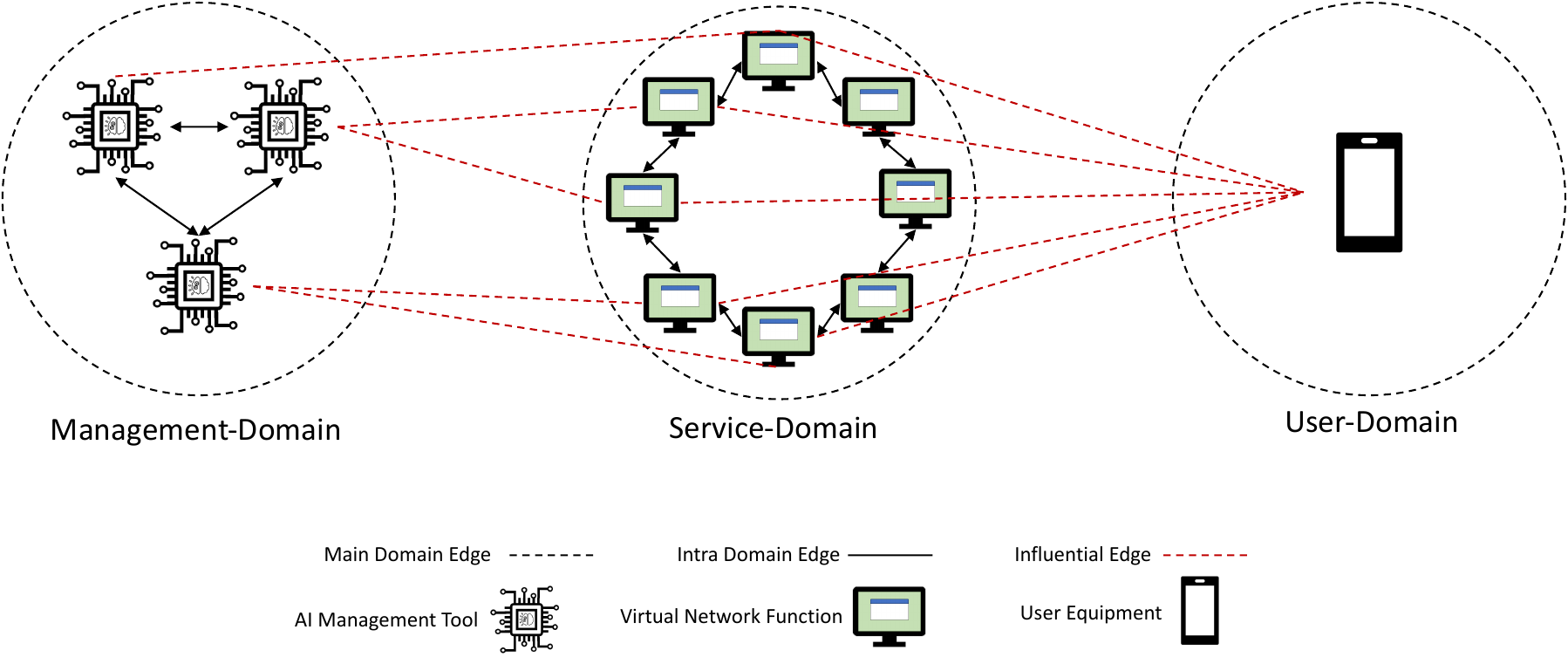}
    \caption{Graphical Modeling of Network.}
    \label{fig:G-net}
\end{figure}

Using Graph Ranking involves modeling elements' influence on one another, using this for an audit system in networking is possible when looking at the different domain levels of the network.  While it follows a top-down approach with the management making changes to the service domain elements that affect the user, the framework can be used in different network configurations.

Figure \ref{fig:G-net} shows how the network can be represented by graphical elements. The links between elements within the same domain are the intra-domain edges while the links between elements that are in different domains are influence edges. The elements within the management domain influence the elements within the service domain, and in turn, influence the user domain. This illustrates the responsibility that the management domain has on the end user. 
This allows for enough of an understanding to model the system. 

The problem being modeled is first finding the modifying agent to the network. In  \ref{eq:alpha}, we start by introducing the variables $a,\mathcal{O}$ as well as a few different sets $N_{Links}, A, N_{Elem}$.  These variables are crucial to identifying the altering agent and are found with $\mathcal{O}$

\begin{equation}\label{eq:alpha}
    a = N_{Links}(A_{\mathcal{O} },N_{Elem}^{\mathcal{O}})
\end{equation}

To find the differences between the new and existing sets of the network elements, we look at the set $N_{Links}$ as in \ref{eq:OEq}. The function then returns the index in which a change was calculated. However, to ensure that the formulation follows realistic constraints, a function such as $\Lambda$.

\begin{equation}\label{eq:OEq}
    \begin{array}{l}
    \mathcal{O} = \Lambda * (\frac{\sum_{\chi \in N_{LinksU}}^{n}(N^{\chi}_{Elem})}{\sum_{\chi \in N_{Links}}^{n}(N^{\chi}_{Elem})}) , \begin{cases} \text{if } \mathcal{O} = 1 , f(\mathcal{O}) = 0\ \\\text{if }\mathcal{O} \neq  1 , f(\mathcal{O}) = n \end{cases}
    \end{array}
\end{equation}

Equation \ref{eq:Lambda} is a binary equation $\Lambda$ to determine if a new allocation of resources is allowed. This allows for other variables to be defined when the constraint has been violated, especially when trying to find the influence of the network elements.

\begin{equation}\label{eq:Lambda}
    \begin{array}{l}
    \Lambda = \frac{\sum_{A \in N_{Elem}^{Res}}^{n}A}{Res_{Total}} \begin{cases}
\text{if }\Lambda = 1 , f(\Lambda) = 1
\\
\text{if }\Lambda \neq 1 , f(\Lambda) = 0

\end{cases}
    \end{array}
\end{equation}

Shown in equation \ref{eq:Rho} is the quantification of the network element's influence on the end user compared to other elements given the element's current resources and the impact on the end user. Now, that the resources allocated by the management agent and the influence on the end user have been found a responsibility can be assigned.

\begin{equation}\label{eq:Rho}
    \rho = \theta * \sum_{\chi \in N_{Elem}}^{n}  N_{res}^{\chi} \times N_{IFactor}^{\chi}
\end{equation}

Equation \ref{eq:Mu} finds the percentage a single network element is responsible for on the end user given the This then allows for responsibility for controlling agents to be established for the change of the networks by comparing the percent of change between network elements and the controlling agents.

\begin{equation}\label{eq:Mu}
    \mu = \frac{\sum_{\chi \in N_Elem}^{n} N_{IFactor}^{\chi}\times N_{Res}^{\chi}}{\sum N_{IFactor}} \times 100
\end{equation}

Equation \ref{eq:nu} first sums all of the percentages that were found by $\mu$, then goes through the set $N_{Links}$ using the subset $A$ to find each network element belonging to the management agent. By dividing the two it can be found how much each agent is responsible for the current state of the network. Then by comparing previous values with current the change in the network and the parties most involved can quickly be determined. 

\begin{equation}\label{eq:nu}
    \begin{array}{l}
    \nu = \frac{1}{\sum_{i=1}^{n} \mu_{i}}\times\sum_{j=1,A\epsilon N_{Links}}^{n}f(\mu)N^{A_{j}}_{Links}
    \\
    \Delta \nu = \nu_{current} - \nu_{prev} 
    \end{array}
\end{equation}

\begin{table}[]
\centering
\resizebox{\columnwidth}{!}{%
\begin{tabular}{|c|c|l|}
\hline
Variable & Description & Range \\ \hline
A & Number of AI agents part of the network &  \\ \hline
N_Elem & Number of VNF elements in the network &  \\ \hline
$N_{Links}$ & List of AI agents and their VNF elements &  \\ \hline
$N_{IFactor}$ & Values of influence that VNF has on end-user &  \\ \hline
$N_{Res}$ & Number of resources VNFs have &  \\ \hline
$\rho$ & Value of influence VNF based on $N_{IFactor}$, $N_{Res}$ &  \\ \hline
$\mathcal{O}$ & Binary operator to detect a change & $0 \leq \mathcal{O} \leq 1$ \\ \hline
$\mu$ & Comprehensive percentage of responsibility &  \\ \hline
$\alpha$ & Altered VNF element with Agent &  \\ \hline
$\nu$ & List of all altered VNF elements &  \\ \hline
$\Lambda$ & Binary operator to detect legal allocation of resources & $0 \leq \Lambda \leq 1$ \\ \hline
\end{tabular}%
}
\caption{List of variables used in problem formulation.}
\label{tab:Vars}
\end{table}

\subsection{Solving using MDP Modeling} \label{MDP}

Realizing what the goal of the RL agent is and the relationships based on graph ranking, an informed decision can be made focusing on the AI tools managing the network; which can correlate to the states of MDP. The actions will be the RL agent guessing based on the observations from the service domain in which the AI tool had made the change. When formulating the MDP the state space can be defined as with three distinct states $S = \left\langle S_{Agent1}, S_{Agent2}, S_{Agent3} \right\rangle$ each representing a management tool.  The reason for this is that when training using the MDP the RL agent must focus on the observations and learn the features of the network, the random action variable is $a_{r}$ following the discrete uniform distribution over the set $[1,2,3]$.The reward will increment if the state and action match. The POMDP allows learning from the observation variables alone.The observation variables are visualized in Figure \ref{fig:G-net} and defined in Table \ref{tab:Vars} and are the influential edges between the nodes across each of the domains, the edges between the management domain and service domain as well as the service domain and the user domain.
\\

\begin{equation}
    r(s, a) =\mathbb{E}[\{^{R_{t} = 1, s = a}_{R_{t}=-1,s\not\equiv a}|S_{t} = s, A_{t}=a]
\end{equation}

 The edges between the management domain and service domain can be denoted as Network Links or $Net_{Links}$. The edges between the service domain and the user domain can be denoted as Network Impact or $Net_{Imp}$. Along with the influential edges, the previous state is added to the observation variable, this will allow learning to take place even when the agent does not provide the correct action. 

\begin{multline}
    \mathcal{O}(s) = (Net^{s}_{obs1}, Net^{s}_{obs2},S_{t-1}|S_{t} = s ,\\
    Net^{s}_{obs1} = Net(s)_{Link}, Net^{s}_{obs2} = Net(s)_{Imp})
\end{multline}

Once the observations are set, the observation function can be defined. The function calculates and gives a value to each of the nodes within the service function determining its weight on the end user. It is then used with the previous state to determine which tool had made the change.

\begin{multline}
    Z = (\mathcal{O},s) = (\frac{s}{\mathcal{O}_{1}}* \mathcal{O}_{2})|S_{t-1}=s,\\
    \mathcal{O}_{1}=Net(s)_{Link}, \mathcal{O}_{2}=Net(s)_{Imp})
\end{multline}

After defining all the functions needed for a POMDP it is now possible to create an environment for a stochastic game to take place and create a simulation. 

\section{Experimentation}\label{Exper}
The experimentation of the system is simulation-based and focused on reproducibility while trying to focus on ease of use. Therefore a tool was developed to achieve both these aspects. The tool creates a testbed that allows the use of different RL algorithms with different simulation parameters, as well as graphical representation and automatic analysis of the simulation. The purpose is for the simulation environment to be customized without having to change source code and RL hyper-parameters. 

\subsection{Simulation Space}


The environment testbed was modeled from the interaction of the network elements' influence on the end user. Each of the network elements has 2 numerical sets, one of the sets representing the allocation of computational resources, and another set representing the influence of the element on the end user. For this simulation, the set representing the influence is not changed during the simulation, while the allocation of the resources is. To simulate an outside management agent making changes to the network, two rolls are performed, one for deciding which agent will make the change, and then a second for increasing or decreasing the resources. The management agents also have a set associated with them containing a list of which network elements they control. While this is taking place within the environment, the influence of each element is being calculated by taking the influence and resource allocation, it is this resulting set that is being shared with the RL agent as observation variables. The goal is from the changing set the RL agent can learn which elements are controlled by which management agent. 

These interactions of the environment are the same across all simulations regardless of the number of network elements or management agents. The tool with which this was integrated allows the user to determine the elements and agents and then randomly rolls the sets of influence and the set of which management agents control which network elements. For reproducibility, these, sets are stored locally and can be loaded in rather than creating new sets each time. For this work, the simulation focused on 3 management tools, with 8 network elements \cite{3gpp_2022}. 


\subsection{Learning Agents}

The same flexibility for the environment wanted to be included in choosing the algorithm of the RL agent, as well as the hyperparameters of the algorithms. For this reason, SB3 and SB3-Contrib\cite{stable-baselines3} were used, to interact with the environment, which includes Augmented Random Search (ARS), Advantage Actor-Critic (A2C), Proximal Policy Optimization (PPO), Quantile Regression Deep Q Network (QRDQN), Recurrent Proximal Policy Optimization (RPPO), and Trust Region Policy Optimization (TRPO). However, a custom leaner can be included however it must follow the input and output of the environment. These algorithms interacted with the environment to train and were evaluated based on the returned rewards. While the training is going the replay buffer from the observation variables is being saved locally to be trained for the ML leaner.

For the ML learner, this was kept constant since the main use is to use the feature importance to determine the specific responsibility of each learner. This model was a deep network with the input correlating to the network elements followed by a hidden layer, and then a softmax layer for the classification of 3 different classes that were determined based on the labels of the data (-25\%, Average, +25\% of the average score). A gradient descent method was used along with standard pre-processing of the data being a 70-15-15 split for training, testing, and validation. The data fed to the ML learner is the replay buffer of the observation variables from the environment during the RL agent runtime. As a checkpoint measure, these observation variables are also stored locally and can be loaded into the tool rather than having to retrain the RL agent.

\subsection{Custom Tool}

As stated a custom tool was developed for this specific experimentation application, where different environments could be run without having to change the source code of the environments. It also allowed for modifying the hyperparameters as well as the algorithms themselves and applying them to a new or existing environment. This was the main motivation when searching for benchmarks, while the actual observation variable's values won't be identical even within the same environment, the constraints of the environment should be kept the same. This also adds scalability to the solution since not every network will have the same number of management agents or network elements, so this allows the solution to be applicable even in those scenarios. The tool utilized Tk-inter a simple Python front-end library and was driven by the custom back-end environments, as well as SB3 for the algorithms. It features a method to load the environment to train the RL agent, as well as a method to load the replay buffer from the simulation to train the ML agent. The framework also handles many different functions to guarantee the success of the models such as data preprocessing and analysis from the ML model. 
For preprocessing of the replay buffer to train the ML, once the buffer is either loaded in or created from the simulation and loaded into the program's memory. For this example, there are 8 network elements, so the observation variables are in arrays size $8 x N$ where $N$ is iterations from the simulations. They first need a label, based on the statistics of the dataset 3 labels are automatically chosen from the average total score of all the elements, where the average is considered normal, a deviation of $-25\%$ is considered poor, and a deviation of $+25\%$ is good. The reason for doing this is so that when the model is trained the network elements impact on these classes can be shown. Once the label is added they are then used to train the ML model. After the model is trained then the framework creates analysis figures in a user-readable format as well as a numerical value for each agent. This value is found by checking the weights for each node that goes to the classification layer as well as the nodes within the classification layer. They are then averaged and then a ratio is found using the total values, shown in equation \ref{eq10}.

\begin{equation}\label{eq10}
\sum_{i=1}^{A}\sum_{n = 1}^{Net_{Elems}\epsilon A_{i}}\frac{Net_{Elems_{n}}}{Total}
\end{equation}

 \begin{figure}[htb]
    \centering
    \includegraphics[width=.48\textwidth]{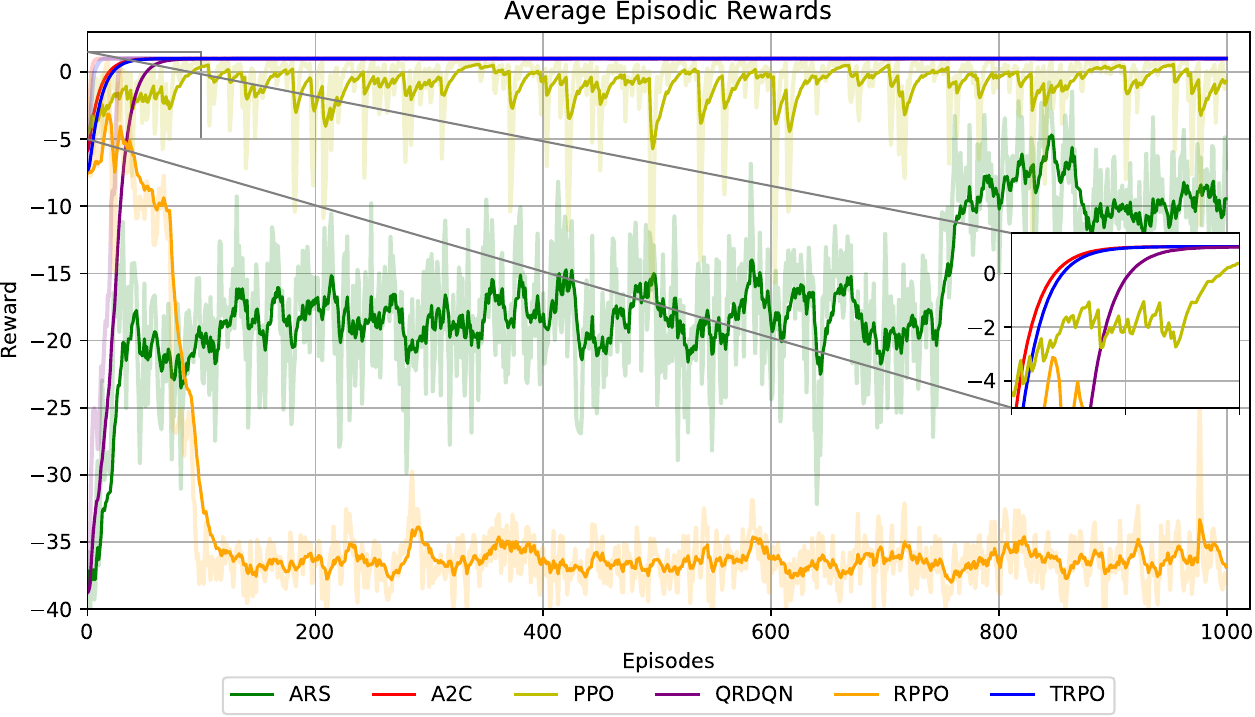}
    \caption{Results from Benchmarks.}
    \label{fig:Reward-BaseLines}
\end{figure}

\section{Results and Analysis}\label{ReA}

\begin{figure}[htb]
    \centering
    \includegraphics[width=.48\textwidth, height=110mm]{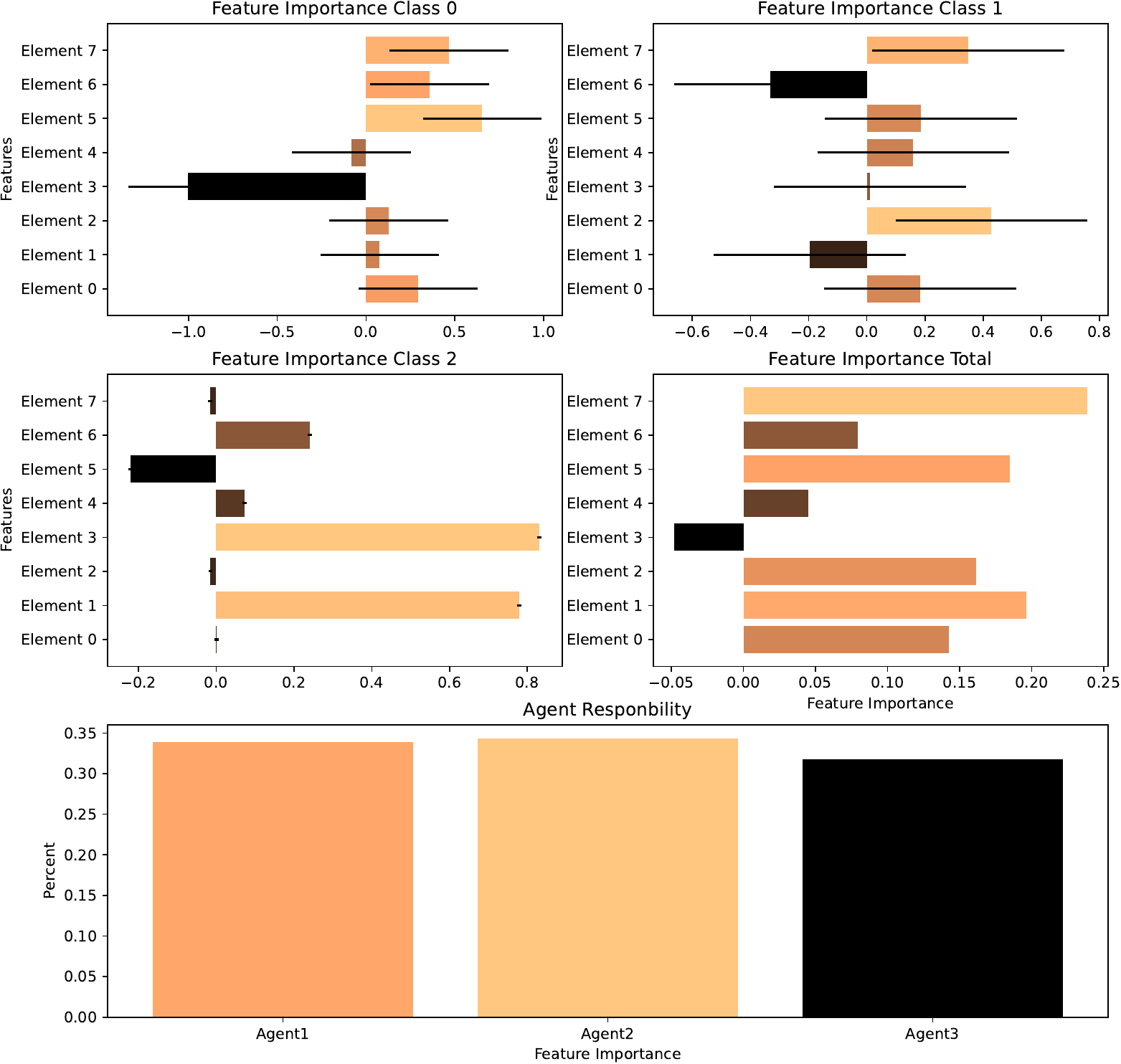}
    \caption{ML Responsibility Quantification.}
    \label{fig:ML}
\end{figure}

Before any training took place a configuration was created with three management agents with eight network elements. The agent controls set was denoted as the following $Control = \begin{Bmatrix}
2 & 4 & 2 \end{Bmatrix}$ meaning Agent 1 has control of 2 network elements, Agent 2 has 4 network elements, and Agent 3 has 2 network elements. The following agent resource allocation is set at $Resource = \begin{Bmatrix}6 & 3 & 1\end{Bmatrix}$. The impact of the network elements $Impact = \begin{Bmatrix}2&1&2&2&1&1&2&2 \end{Bmatrix}$. This configuration was saved was used for all of the testing. Once this configuration was created benchmarks were collected using default hyperparameters for each RL learning algorithm, shown in \ref{fig:Reward-BaseLines}. Keeping in mind that the learners are operating in a POMDP space rather than a traditional MDP space, this is why each was tested to see the performance in a custom environment. The optimal reward is one, showing that the algorithm could successfully determine the change in the environment to the management agent during a single time step. Figure \ref{fig:Reward-BaseLines}, shows the different algorithms and shows that A2C, TRPO, and QRDQN outperform the other methods. Some of the other methods are more advanced such as PPO, RPPO, and ARS do not perform as well. For all of the algorithms the default hyperparameters were used, this included learning rate, gamma, and steps. The episodes were set to 1000 as an arbitrary stopping value and were kept due to the ARS algorithm showing improvement later on. However, the other algorithms flatten out at around 200 episodes. The methods that do perform well combine value-based learning with input and output rewards. The reason for these algorithms performing better may be that the environment's constraints are easily learnable given the observation variables. When the best performers were tested, they returned the following accuracies out of 100 test scenarios, TRPO 96\%, A2C 89\%, and QRDQN with 93\%. While this testing for the RL was taking place the replay buffer of the observation variables was being saved to be used as input data for the ML model. Training the model was simple with an arbitrary stopping point selected at 100 iterations, this was because the data inputted consisted mainly of integers and not complex data, so learning past this point would result in overfitting. While being a deep neural network it is not complex enough to weigh a significant computational load on modern processors. With this being said contribution this system is for the feature importance so extracting the weights and biases from the system is crucial. To be able to access the weights directly the neural net was built from scratch using Numpy arrays and standard operations. Gradient descent was applied as the learning method, where the inputs correlated to the network elements and activation functions used are ReLu, except for the final layer utilizing a Softmax.

Figure \ref{fig:ML} shows how the ML model's weights are used to quantify the responsibility. Each input element's weight on the classifier is shown after training, and the black line(bias) shows the degree of error. This is done for each of the classification nodes that correlate to the labels given, therefore three figures. The total feature importance is then found and represented on the following graph. Given the simulation parameters, the framework can match the network elements to each of the agents, this is done rather than using the prediction from the RL agent to prevent mistakes on the graphs. From the total, the ratios are calculated to give a comprehensive percentage of which agent is responsible for each agent. In this simulation, Agent 1 has a responsibility of $33.9\%$, Agent 2 has a responsibility of $34.3\%$, and Agent 3 has a responsibility of $31.81\%$. These values can change given different simulation parameters. 

\section{Conclusion and Future Research Directions}\label{Conc}

This paper presented a framework of DRL and ML models to help identify and quantify responsibility in a multi-operator managed network. The system model presented formulates the problem through the influence each elements have on one another. From this formulation, a custom scalable environment was created to train both DRL models and ML models. The DRL models took advantage of modeling through POMDP and the ML model used the replay buffer from this simulation. This framework also features an analysis section to identify responsibilities from the training, and it was all incorporated in an ease-of-use tool. The contribution is to close the gap of responsibility for AI management tools in a network and to ensure their fair and ethical use. 

Further contributions to this work can include many different solutions while following the main objective. A different environment approach can be taken where an adversarial game where rather than a single agent making actions and the network following a stochastic reaction, a multi-agent game can take place. This can follow an approach similar to Game Theory. Also, further improvements can be made to the prototype tool by adding support for more different and custom algorithms can be implemented. An automatic hyperparameter can also be implemented into the tool to help decide the best options for the custom environment.

\bibliographystyle{IEEEtran}
\bibliography{refernces1}
\end{document}